\documentclass[letterpaper]{article} 
\usepackage{aaai2026}  
\usepackage{times}  
\usepackage{helvet}  
\usepackage{courier}  
\usepackage[hyphens]{url}  
\usepackage{graphicx} 
\usepackage{natbib}  
\usepackage{caption} 
\usepackage{algorithm}
\usepackage{algorithmic}

\usepackage{booktabs}  
\usepackage{array}   
\usepackage{multirow}
\usepackage{amsmath}
\usepackage[table]{xcolor}

\title{Spatial-Morphological Modeling for Multi-Attribute Imputation of Urban Blocks}

\author {
    Vasilii Starikov\textsuperscript{\rm 1},
    Ruslan Kozliak\textsuperscript{\rm 1},
    Georgii Kontsevik\textsuperscript{\rm 1},
    Sergey Mityagin\textsuperscript{\rm 1}
}
\affiliations {
    \textsuperscript{\rm 1}ITMO University, St. Petersburg, Russia\\
    mityagin@itmo.ru 
}

\begin{document}

\maketitle

\begin{abstract}

Accurate reconstruction of missing morphological indicators of a city is crucial for urban planning and data-driven analysis. This study presents the spatial-morphological (SM) imputer tool, which combines data-driven morphological clustering with neighborhood-based methods to reconstruct missing values of the floor space index (FSI) and ground space index (GSI) at the city block level, inspired by the SpaceMatrix framework. This approach combines city-scale morphological patterns as global priors with local spatial information for context-dependent interpolation. The evaluation shows that while SM alone captures meaningful morphological structure, its combination with inverse distance weighting (IDW) or spatial k-nearest neighbor (sKNN) methods provides superior performance compared to existing SOTA models. Composite methods demonstrate the complementary advantages of combining morphological and spatial approaches.

\end{abstract}


\section{Introduction}

Today's world is developing rapidly, changing its appearance under the influence of new technologies. Cities are changing physically and socially under the influence of new technologies, offering new social and economic models of behavior, which leads to the emergence of new ways of using land resources \cite{xia2025fundamental}. Advances in telecommunications allow people to work remotely, potentially reducing the need for traditional office space. Autonomous vehicles eliminate the need for parking spaces, opening up opportunities for the creation of recreational areas and other public facilities \cite{nourinejad2018designing}. 

AI is influencing the functional spaces of cities, providing new spatial representations of (de)centralization processes and assessing the implementation of new forms of urban mobility on spatial equality. The results of such changes can be seen at the neighborhood level: changes in the structure of the layout and recombination of urban space functions can lead to the modular organization of small neighborhoods and the flattening of the street system \cite{villegas2025design}, \cite{yu2025can}.

The emergence of foundation models and large language models (LLMs) represents a transformative shift in urban data science, offering new paradigms for addressing data incompleteness and spatial analysis challenges. Urban Foundation Models (UFMs)—large-scale models pre-trained on vast amounts of multi-source, multi-granularity urban data—have demonstrated capability in capturing fundamental patterns across macro-level city systems, meso-level urban structure, and micro-level neighborhood dynamics \cite{chen2025spatialllm} \cite{zhang2025urbangeneralintelligencereview}. Recent applications have extended LLMs to urban geospatial intelligence through zero-shot analysis, enabling natural language interactions with geographic databases and supporting complex spatial queries without domain-specific training. Generative AI approaches have been applied to automated land-use configuration, learning optimal spatial arrangements from heterogeneous datasets including mobility patterns, POI distributions, and environmental factors \cite{wang2025generativeaimeetsfuture}. However, while these foundation models excel at reproducing abstract theoretical forms and generating plausible urban patterns, they face significant limitations in capturing statistical complexity and parametric accuracy of real-world urban systems \cite{zhang2025genaimodelscaptureurban}. Moreover, their application to fine-grained morphological reconstruction tasks—such as imputing specific built-form attributes at the block level—remains underexplored, particularly regarding the integration of local spatial context with global morphological knowledge. This gap motivates approaches that combine morphological clustering with local features to achieve context-aware, spatially coherent predictions grounded in empirical urban form patterns

Urban data are often incomplete or inconsistent, particularly for built-form attributes such as floor space, footprint area, and land-use composition. 
Accurate reconstruction of these features is critical for practical urban planning and territorial management. For example when new development areas are introduced or existing urban zones are modified according to a master plan, planners need quantitative predictions of likely building density and coverage. 
This information supports decision-making by bridging the gap between zoning regulations and expected morphological outcomes.

This work investigates whether incorporating global morphological knowledge of an urban area can enhance the accuracy and spatial coherence of imputed local built-form parameters for individual urban blocks.

This study introduces a method that combines morphological clustering with local spatial features to impute missing built-form attributes, supporting accurate, context-aware urban modeling and planning.

\section{Related Work}

\subsection{The problem of incomplete and uneven urban data}

When working with digital twins and urban models, it is critical to have a complete set of semantic and geometric attributes (height, number of floors, type of use, density, land-use indicators, etc.). In practice, data is often fragmented: some features are only available locally, while others are completely absent, making the task of automatic restoration (imputation/inpainting/prediction) crucial for applications in planning, energy, traffic flow modeling, etc. Three main approaches to this problem are emerging in the literature: image-based methods, methods using mobile and behavioral data, and data-driven methods based on digital twins and graph representations.

\subsection{Image-based methods}

One of the dominant approaches is the use of satellite and street-level images to reconstruct the morphology and semantic attributes of buildings (contours, roofs, facades, number of floors, height) using classical methods (shadow dependencies, stereo, LiDAR) and modern neural network architectures ~\cite{li2012optimisation,klonus2011image,srivastava2018deep,lienhard2020automatic}. 

The integration of images and spatial context through graph models has shown an improvement in prediction quality: combining street images with graph neural networks (GraphSAGE) improves the accuracy of building attribute predictions compared to purely visual approaches~\cite{lei2024predicting}. However, image-based methods are limited by coverage (not all areas have street-view or high-quality images), are sensitive to weather and temporal factors, and require costly preprocessing~\cite{zhang2021cross}.

\subsection{Methods based on mobile and behavioral data}

Another large group of studies uses mobile operator data, GPS tracks, check-ins, and POIs to assess the functional purpose of areas and identify temporal patterns of activity. Studies show that mobility data correlates well with land use and allows for effective classification of areas according to their functionality~\cite{lenormand2015comparing,frias2013urban}. However, these sources are often closed and subject to legal and privacy restrictions, which hinders their large-scale use in open digital twins.

\subsection{The role of land use and causal approaches}

Research on land use~\cite{lenormand2015comparing} emphasizes that the spatial organization of land use types is closely related to the morphology and dynamics of a city—the distribution of land use correlates with density, transport connectivity, and the shape of neighborhoods. Moreover, causal machine learning methods have shown that the influence of land use on building height is heterogeneous and spatially non-uniform: the same type of land use can have different effects in the city center and on the periphery~\cite{chen2024inferring}. This reinforces the idea that land use is not just a useful correlate, but also a causal factor to consider when modeling and restoring morphological features.

\subsection{Data-driven approaches and digital twins; the role of OSM}

Data-driven approaches based on digital twins (2D/3D models, OpenStreetMap, etc.) provide a reproducible and scalable basis for attribute recovery. It has been shown that OpenStreetMap (OSM) data combined with geometric and contextual features can successfully predict building types and enrich OSM tags~\cite{atwal2022predicting,schwarz2023classification}. These approaches are advantageous in terms of data availability and experiment replicability.

\subsection{Digital twins of cities}

The concept of digital twins is becoming a key technology for creating smart and sustainable cities. A digital twin of a city is a virtual replica of a physical city that integrates data from various sources to support planning, management, and decision-making~\cite{batty2018digital,dembski2020urban}.

Cities around the world are developing their digital twins. For example, the city of Zurich has created a digital twin to support urban planning, including 3D models of buildings, street networks, mobility simulations, and urban climate analysis~\cite{schrotter2020digital}. The city of Herrenberg in Germany uses a digital twin for participatory urban planning, integrating BIM and GIS data with sensor networks and volunteered geographic information~\cite{dembski2020urban}.

It has been highlighted that integrating GIS and BIM is important for creating effective digital twins of cities~\cite{xia2022study}. GIS provides geospatial data and spatial analysis at the city level, while BIM provides detailed information about buildings at the micro level. The ontological approach to data integration is recognized as the most promising for achieving system interoperability.

Digital twins enable predictive modeling and simulation of city development scenarios. ~\cite{schrotter2020digital} describe the use of Zurich's digital twin to analyze the urban climate and heat islands, assess the energy efficiency of buildings, plan high-rise buildings, manage the urban cadastre, and conduct architectural competitions using AR/VR technologies.

An important aspect is data openness. Zurich publishes its 3D geodata under an Open Government Data license, which promotes the development of innovative applications and the democratization of urban data~\cite{schrotter2020digital}.

The social aspects of digital twins have been emphasized, showing that their use in virtual reality can significantly improve citizen participation in urban planning~\cite{dembski2020urban}. A survey of 39 participants indicated that digital twins are perceived as interesting (2.54 out of 3.00), understandable (2.37), and concrete (2.14).

Digital twins are also used to manage historical heritage, simulate floods and emergencies, optimize construction processes, and manage infrastructure~\cite{xia2022study}. The integration of real-time IoT sensors transforms static models into dynamic systems capable of responding to changes in the urban environment.

\subsection{Graph Neural Networks (GNN) for Urban Tasks}

Graph representations (nodes --- buildings, blocks; edges --- adjacency, transport links, spatial proximity) and GNNs proved particularly effective for accounting for neighborhood influences when restoring features. Classic architectures --- ChebNet, GraphSAGE, GAT, Graph U-Nets, etc. --- have proven their ability to aggregate neighbor features and extract structural patterns~\cite{defferrard2016convolutional,hamilton2017inductive,velickovic2018graph,gao2019graph}. 

In applied research, GraphSAGE has been used to predict building characteristics on a city-wide scale, while ChebNet, Graph U-Nets, and attention-based mechanisms similar to GAT have been applied for hierarchical aggregation and weighting neighbor contributions~\cite{lei2024predicting,liu2025urban}. These techniques have demonstrated high effectiveness for classifying street network morphological patterns and predicting neighborhood attributes, justifying the use of GNNs as the main architecture for multi-attribute imputation at the neighborhood level.

\subsection{Multi-view imputation of urban data}

A specific subgroup of methods focuses on restoring tabular urban characteristics collected region-by-region. The SMV-NMF/SMKC-AWNMF approach combines spatial multi-kernel clustering and adaptive-weight non-negative matrix factorization to recover multi-species statistical data; the method showed high accuracy on several real-world datasets and good transferability between cities~\cite{gong2022spatial}.

A follow-up study formalized Spatial Correlation Learning by combining spatial multi-kernel clustering and adaptive-weight NMF within SMV-NMF and demonstrated superiority over KNN, Bayesian PCA, and similar baseline methods on six real-world datasets~\cite{gong2023missing}. The key conclusion is that spatial structure and multi-view interactions are critical for successful imputation of urban tables; this complements and partially overlaps with the idea of graph methods, which explicitly use neighborhood through graph topology.

\subsection{Comparison of levels: building vs neighborhood; multi-attribute imputation}

Most studies focus on predicting individual attributes of individual buildings (height, number of floors, type). However, many planning tasks require aggregated and simultaneously multi-parameter information at the neighborhood (block) level. Our approach deliberately shifts the focus to neighborhoods as graph nodes and solves the problem of simultaneous restoration of multiple attributes (multi-output imputation), taking into account neighborhood geometry, neighborhood connections, and land-use characteristics. 

Ideas on multi-view spatial imputation ~\cite{gong2022spatial,gong2023missing} and the causal influence of land use ~\cite{chen2024inferring} reinforce the justification for including both spatial correlation and functional purpose features in the model; Other works ~\cite{lei2024predicting,liu2025urban} show that GNN architectures provide a practical tool for implementing this idea.

\subsection{Morphological Clustering and SpaceMatrix}

SpaceMatrix is a framework for classifying urban blocks into morphological types based on built-form indices such as FSI (Floor Space Index) and GSI (Ground Space Index)~\cite{wu2024spacematrix, berghauser2018typology}. 
It uses unsupervised clustering (e.g., KMeans) on normalized morphological features to derive characteristic urban forms, which can then guide analysis of urban patterns, functional zoning, or reconstruction of missing data. This approach has been applied successfully for morphological assessment and comparative studies of urban neighborhoods, providing a scalable and interpretable representation of city structure.

\subsection{Conclusion}

A comparison of approaches shows that image-based methods provide visual cues but suffer from coverage and preprocessing issues; mobile data offer dynamic information but have limited availability. 
Tabular imputation methods, such as SMV-NMF, are effective for multi-view urban data but may overlook morphological regularities. 

Morphological clustering methods, such as SpaceMatrix, classify urban blocks into characteristic types based on built-form indices (e.g., FSI, GSI)~\cite{wu2024spacematrix, berghauser2018typology}. 
These clusters capture meaningful patterns in urban morphology and can serve as a basis for imputing missing block-level attributes. 
While prior studies typically apply clustering alone, it is plausible that combining morphological clusters with spatially aware techniques (e.g., neighborhood averaging or SKNN) could enhance reconstruction accuracy, providing both local and global context in urban data recovery.




\section{Problem Formulation}

Urban blocks exhibit complex relationships between land-use composition and built-form intensity.  
In this study, each block $b_i$ is represented by a feature vector  
\[
X_i = \{x^{(r)}, x^{(rec)}, x^{(bus)}, x^{(ind)}, x^{(tr)}, x^{(sp)}, x^{(agr)}, s_i\},
\]
where $x^{(k)} \in [0,1]$ denote fractional land-use shares for residential, recreational, business, industrial, transport, special, and agricultural functions, and $s_i$ is the total site area (m$^2$).  

The target built-form attributes are expressed in normalized form as  
\[
Y_i = \{\text{FSI}_i, \text{GSI}_i\},
\]
where the \textit{floor space index} (FSI) and the \textit{ground space index} (GSI) represent the ratio of total floor and footprint area to site area, respectively.  
These indicators capture the intensity and compactness of development, making them invariant to block size and more consistent across different land-use types.  

The imputation task is formulated as learning a mapping
\[
\hat{Y}_i = f(X_i),
\]
which predicts the built-form indicators of block $i$ based on its land-use composition and spatial characteristics.  
Unlike conventional imputers that rely on spatial proximity or local averaging, the proposed approach introduces a \textit{morphological regularization} step: each block is first associated with a probabilistic morphological cluster (or \textit{urban form type}) inferred from the joint FSI–GSI distribution.  
Then, the characteristic built-form profile of the most probable cluster is used to reconstruct the missing values.  
This design captures global regularities of urban morphology, enabling consistent and interpretable recovery of built-form indicators even without explicit neighborhood information.

\paragraph{Empirical motivation.}
Preliminary statistical analysis confirmed both land-use–dependent variation and spatial autocorrelation in the studied data. 
Each urban block was assigned to a \textit{dominant land-use group} corresponding to the category with the highest proportional share among residential, recreational, business, industrial, transport, special, and agricultural uses.  
A one-way ANOVA revealed statistically significant differences in built-form intensity across these groups 
($F=288.1$, $p<0.001$ for FSI; $F=392.0$, $p<0.001$ for GSI), 
indicating that functional composition strongly influences the compactness and density of development.  
Moran’s I statistics further demonstrated pronounced spatial autocorrelation 
($I=0.53$, $p=0.001$ for FSI; $I=0.57$, $p=0.001$ for GSI), 
suggesting that similar morphological patterns cluster spatially across the city.  
Finally, Pearson’s correlations ($r=0.77$ between FSI and GSI, and $r=-0.06$ to $-0.10$ with site area) 
indicate a strong internal coupling between built-form indicators but minimal dependence on block size.  
Together, these findings support the need for an imputation strategy that incorporates both morphological structure and spatial regularity of urban form.
formed imputation strategy that accounts for both morphological and locational dependencies.

\section{Methodology}

\subsection{Spatial-morphological Imputation Model}

The proposed model estimates built-form indicators (FSI and GSI) for urban blocks with missing values. The method relies on the idea that each block can be associated with a \textit{morphological type} defined by the joint distribution of FSI and GSI.

First, blocks with valid data are standardized and clustered in the $(\text{FSI}, \text{GSI})$ space using K-means. Each cluster $c_k$ represents a characteristic morphological pattern (e.g., compact mid-rise, low-density), and its centroid provides typical FSI and GSI values.

Next, a CatBoost classifier predicts the probability distribution over these clusters for each block based on its land-use composition and site area:
\[
P(c_k \mid X_i) = \text{CatBoost}(X_i),
\]
where $X_i$ includes the shares of residential, recreational, business, industrial, transport, special, and agricultural uses, as well as total site area.
The final FSI and GSI estimates are obtained as probability-weighted averages of cluster-level medians:
\[
\hat{\text{FSI}}_i = \sum_k P(c_k \mid X_i)\,\text{FSI}_k, \quad
\hat{\text{GSI}}_i = \sum_k P(c_k \mid X_i)\,\text{GSI}_k.
\]
This formulation captures city-wide morphological regularities while remaining interpretable and spatially coherent.

\subsection{Model Overview and Workflow}

\begin{figure*}[t]
    \centering
    \includegraphics[width=\linewidth]{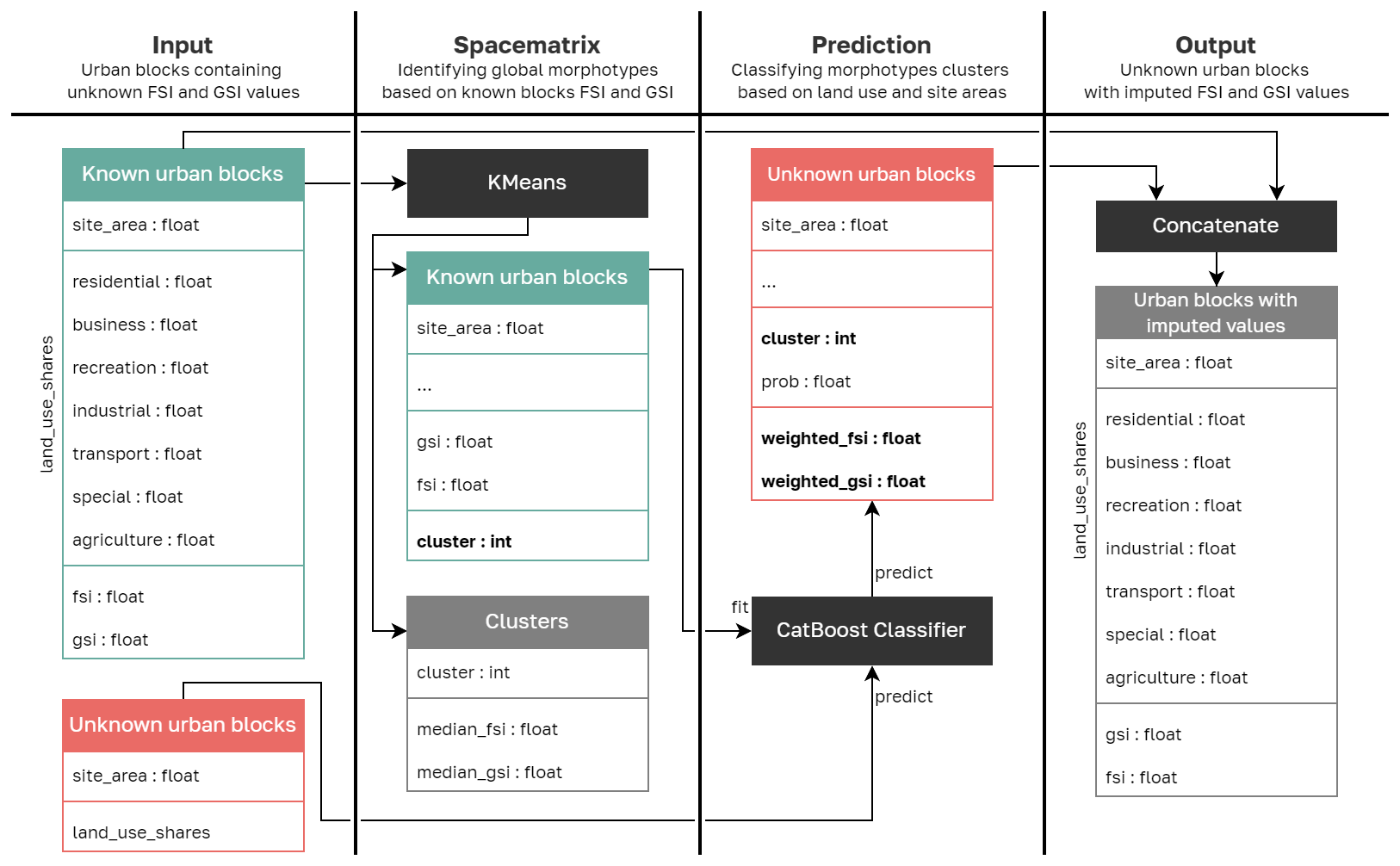}
    \caption{Workflow of the Spatial-morphological (SM) imputation model. 
    The model derives morphological clusters from observed data and predicts missing built-form indicators based on land-use composition and site area.}
    \label{fig:workflow}
\end{figure*}

Figure~\ref{fig:workflow} illustrates the workflow of the Spatial-Morphological (SM) imputation model.  
The process consists of four stages:

\begin{enumerate}
    \item \textbf{Input:} compilation of block-level data, including:
    \begin{itemize}
        \item \textit{site\_area} — total block area;
        \item \textit{land-use proportions} — shares of residential, business, recreational, industrial, transport, special, and agricultural uses;
        \item unknown \textit{FSI} and \textit{GSI} values to be imputed.
    \end{itemize}
    
    \item \textbf{Spacematrix (Morphological clustering):} using only blocks with known FSI and GSI:
    \begin{itemize}
        \item cluster blocks in the standardized $(\text{FSI}, \text{GSI})$ space using K-means;
        \item assign cluster IDs to known blocks;
        \item compute median FSI and GSI for each cluster.
    \end{itemize}
    
    \item \textbf{Prediction (Cluster assignment for unknown blocks):} 
    \begin{itemize}
        \item train a CatBoost classifier on known blocks to predict cluster probabilities based on land-use shares and site area;
        \item for each unknown block, compute predicted cluster ID, probability, and probability-weighted FSI/GSI values.
    \end{itemize}
    
    \item \textbf{Output (Imputed values):} generate a table of blocks with missing FSI and GSI, now imputed using probability-weighted cluster medians.
\end{enumerate}

\section{Data}

Urban block geometries and topological relationships were initially generated using the \textbf{BlocksNet} framework\footnote{\url{https://github.com/aimclub/blocksnet}}, which provides tools for urban blocks layer generation, spatial aggregation, and morphometric analysis. 
The generated layer was subsequently reviewed and manually refined to correct topological inconsistencies. 
Building-level and land-use-level attributes were then aggregated to the block scale using BlocksNet’s tools.

For each urban block, the following features were extracted and aggregated:

\begin{itemize}
    \item \textbf{Land-use composition:} fractional shares of residential, business, recreational, industrial, transport, special, and agricultural functions;
    \item \textbf{Geometric attributes:} site area (m$^2$);
    \item \textbf{Built-form indices:} floor space index (FSI) and ground space index (GSI), defined as:
    \begin{align*}
    \text{FSI} &= \frac{\text{Total floor area of buildings}}{\text{Site area}}, \\
    \text{GSI} &= \frac{\text{Total building footprint area}}{\text{Site area}}.
    \end{align*}
\end{itemize}

A descriptive summary of these features is presented in Table~\ref{tab:data_stats}. 
To evaluate imputation methods, missing values in FSI and GSI were introduced synthetically, allowing controlled comparisons of SM, SKNN, IDW, SMV-NMF, and their hybrid combinations.

\begin{table}[t]
\centering
\caption{Descriptive statistics of urban block features (Saint Petersburg).}
\label{tab:data_stats}
\begin{tabular}{lccc}
\toprule
Feature & Mean & Std & Range \\
\midrule
Residential share & 0.478 & 0.435 & [0, 1] \\
Business share & 0.031 & 0.146 & [0, 1.00] \\
Recreation share & 0.198 & 0.340 & [0, 1.08] \\
Industrial share & 0.048 & 0.188 & [0, 1.00] \\
Transport share & 0.190 & 0.301 & [0, 1.00] \\
Special share & 0.013 & 0.103 & [0, 1.00] \\
Agricultural share & 0.014 & 0.107 & [0, 1.00] \\
Site area (m$^2$) & 1.46e5 & 3.67e5 & [2.71e-20, 1.40e7] \\
FSI & 0.377 & 0.631 & [0, 9.15] \\
GSI & 0.110 & 0.134 & [0, 1.10] \\
\bottomrule
\end{tabular}
\end{table}

Some blocks exhibit extreme values due to data inconsistencies or overlapping functional zones, resulting in GSI or land-use shares slightly exceeding 1. 
These outliers are retained for completeness but do not significantly affect overall imputation results.

\section{Experiments}

We evaluate the proposed \textbf{Spatial-morphological Imputer} (SM-Imputer, hereafter \textbf{SM}) on the Saint Petersburg urban block dataset.
To systematically assess imputation performance, we introduced synthetic missing values in \texttt{FSI} and \texttt{GSI} at varying missing rates from 10\% to 70\%. 
For each missing rate, a common mask was applied across all imputation methods to ensure fair comparison. 
Each experiment was repeated with 100 independent random masks, and the resulting metrics were averaged.

\subsection{Baseline Methods}

The baselines are described concisely as follows~\cite{gong2022spatial}:

\begin{itemize}
    \item \textbf{IDW (Inverse Distance Weighting):} Distance-weighted average of neighboring observations. IDW is a standard spatial interpolation method widely used in urban analytics and geographic applications, providing a simple reference for local spatial dependence in built-form attributes.
    
    \item \textbf{sKNN:} Spatial $k$-nearest neighbors using block centroids. sKNN captures the influence of nearby blocks while allowing multi-feature imputation, making it directly relevant for urban block-level data where neighborhood similarity is expected.
    
    \item \textbf{SMV-NMF:} Spatially-informed multi-view non-negative matrix factorization. SMV-NMF is a recent multi-view imputation method designed for tabular urban data, combining spatial correlations with multiple attribute types, which aligns closely with the multi-attribute nature of the dataset.
\end{itemize}

Our method, \textbf{SM}, estimates missing FSI and GSI by predicting the most likely morphological cluster for each block, as described in the methodology section. In addition, we explore \textit{hybrid approaches}, where SM predictions are combined with outputs from IDW, sKNN, or SMV-NMF, which often provide complementary improvements by balancing local spatial effects and global morphological patterns.

\subsection{Evaluation Metrics}

Imputation performance was quantified using mean absolute error (MAE), root mean squared error (RMSE), coefficient of determination ($R^2$), and robust $R^2$ ($R^2_{robust}$). 
Metrics were averaged over the 100 masks for each missing rate. 
The following section presents the results, highlighting the advantages of SM and its hybrid combinations relative to baseline methods.

\subsection{Results}

Table~\ref{tab:results} reports the averaged imputation performance across all missing rates for both individual and composite methods. 
The base \textbf{SM} method, which predicts missing FSI and GSI using morphological clusters alone, shows moderate accuracy, outperforming SMV-NMF but not IDW or sKNN on its own. 
However, combining SM with local neighborhood-based imputers yields a significant improvement: both \textbf{SM + IDW} and \textbf{SM + sKNN} achieve the lowest MAE and RMSE, with the highest $R^2$ values.

\begin{table}[t]
\centering
\caption{Imputation performance on Saint Petersburg urban blocks (averaged over 100 masks per missing rate).}
\label{tab:results}
\begin{tabular}{lcccc}
\toprule
Method & MAE & RMSE & $R^2$ & $R^2_{robust}$ \\
\midrule
IDW & 0.1713 & 0.3085 & 0.3583 & 0.8859 \\
sKNN & 0.1731 & 0.2991 & 0.4002 & 0.8603 \\
SM & 0.1837 & 0.3142 & 0.3409 & 0.8556 \\
SM + IDW & 0.1629 & 0.2733 & 0.5038 & 0.8461 \\
SM + sKNN & 0.1647 & 0.2715 & 0.5120 & 0.8285 \\
SM + SMV-NMF & 0.2203 & 0.3291 & 0.2708 & 0.5987 \\
SMV-NMF & 0.2718 & 0.3769 & 0.0264 & 0.1668 \\
\bottomrule
\end{tabular}
\end{table}

\begin{figure*}[t]
    \centering
    \includegraphics[width=0.9\linewidth]{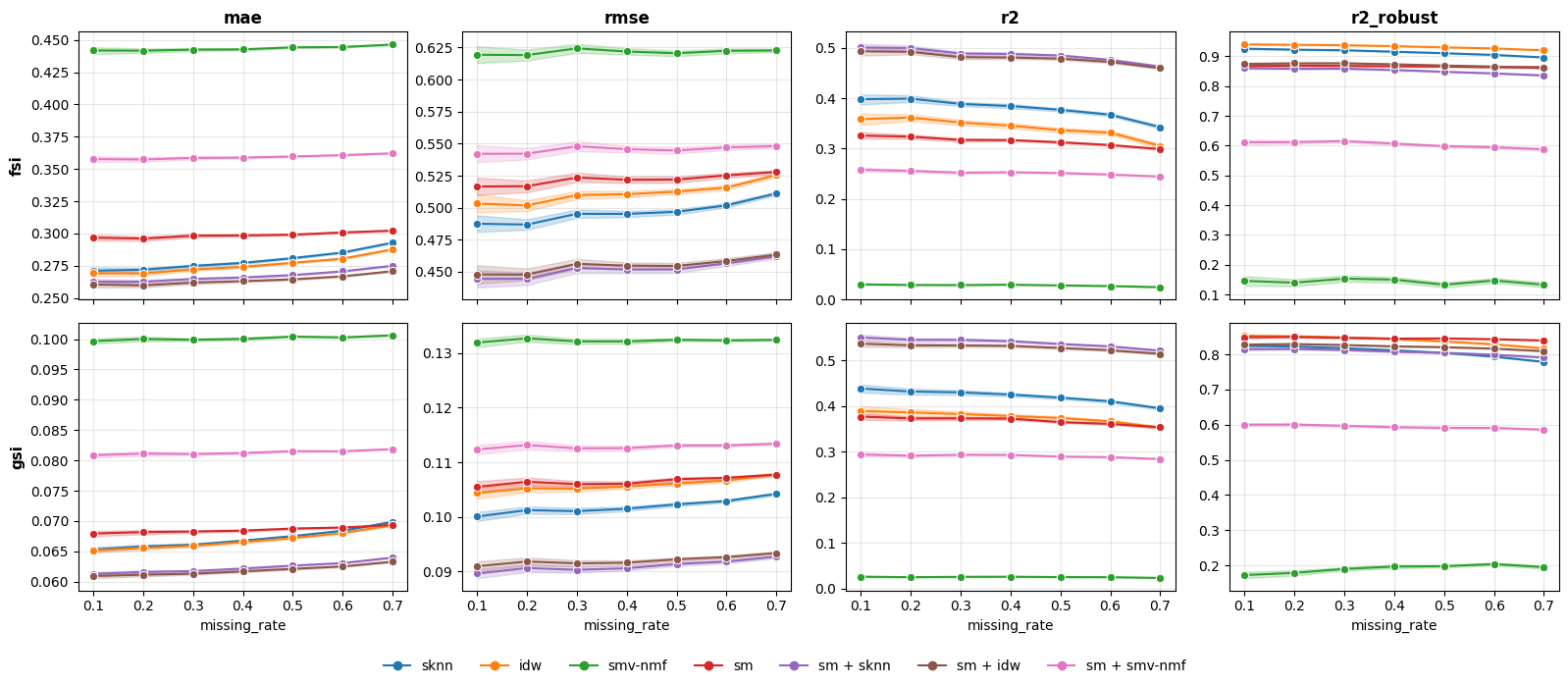}
    \caption{Performance of SM, baseline methods, and composite imputers across missing rates for each reconstructed feature. 
    Composite methods (SM + IDW, SM + sKNN) achieve the best results.}
    \label{fig:metrics_vs_missing}
\end{figure*}

Figure~\ref{fig:metrics_vs_missing} visualizes the performance trends across missing rates for each reconstructed feature. 
Composite methods (\textbf{SM + IDW} and \textbf{SM + sKNN}) consistently outperform individual imputers, highlighting the complementary strengths of morphological and local neighborhood information. 

The dataset and code used in these experiments are publicly available\footnote{\url{https://anonymous.4open.science/r/smi}}.

\section{Discussion}

The experimental results indicate that the \textbf{Spatial-morphological imputer} captures meaningful morphological patterns in urban blocks, even without explicit use of local neighbors. 
While the base SM method alone provides moderate accuracy, its combination with neighborhood-based imputers (IDW or sKNN) leads to substantial improvement, achieving the lowest MAE and RMSE across all missing rates. 
This suggests that local spatial information and global morphological priors are complementary: one tends to slightly underestimate (or overestimate) values while the other compensates, yielding more balanced predictions.

The strongest relative gains are observed at higher missing rates, when purely local methods suffer from sparse information. 
In such scenarios, SM-derived morphological priors offer a robust global structure, enabling more accurate and spatially coherent imputation. 
These findings highlight the practical value of integrating morphological knowledge with traditional spatial imputation techniques for urban datasets with incomplete or noisy measurements.

Unlike methods relying on predefined morphological typologies, SM dynamically derives morphological clusters from available data, allowing adaptation to the specific urban fabric of the city under study. 
Future work could explore end-to-end learning of morphological representations, or extend the framework with additional sources of spatial connectivity (e.g., transport networks or functional adjacency) to further improve imputation robustness and interpretability.

\section{Conclusions}

This study introduced the \textbf{Spatial-morphological (SM) imputer}, a framework that leverages data-driven morphological clustering to reconstruct missing FSI and GSI values at the urban-block level. 
Evaluation on the Saint Petersburg dataset demonstrated that SM alone captures meaningful morphological structure, while its combination with neighborhood-based methods (IDW or sKNN) consistently achieves the highest accuracy, highlighting the complementarity of global morphological priors and local spatial information.

The results underscore the value of integrating city-scale morphological patterns with spatial context for robust, context-aware urban data reconstruction. 
Future work will focus on extending the approach to multi-city datasets, exploring transferability across different urban morphologies, and integrating SM into broader pipelines for generative urban modeling, simulation, and planning applications.

\vspace{.2em}

\section{Acknowledgments}

\bigskip
\noindent This work supported by the Ministry of Economic Development of the Russian Federation (IGK 000000C313925P4C0002), agreement No139-15-2025-010.

\bibliography{aaai2026}

\end{document}